\documentclass[runningheads]{llncs}
\usepackage{todonotes}
\usepackage{graphicx}
\usepackage{wrapfig}
\usepackage[hidelinks,colorlinks=true,urlcolor=blue,linkcolor=black,citecolor=black]{hyperref}
\usepackage{url}
\usepackage{glossaries}
\usepackage{float}
\glsdisablehyper

\newglossaryentry{s4n}
{
    name=STAMP 4 NLP,
    description={STAMP 4 NLP}
}

\begin{document}
\title{STAMP 4 NLP -- An Agile Framework for Rapid Quality-Driven NLP Applications Development}
\titlerunning{STAMP 4 NLP -- Agile NLP Application Development}
%
\author{Philipp Kohl\inst{1} \and
Oliver Schmidts\inst{1} \and
Lars Klöser\inst{1} \and
Henri Werth\inst{1} \and
Bodo Kraft\inst{1} \and
Albert Zündorf\inst{2}}

\institute{FH Aachen - University of Applied Sciences, 52428 Jülich, Germany \and University of Kassel, 34127 Kassel, Germany \email{\{p.kohl,schmidts,kloeser,werth,kraft\}@fh-aachen.de}\\ \email{zuendorf@uni-kassel.de}}

\authorrunning{P. Kohl et al.}
\maketitle

\begin{abstract}

The progress in natural language processing (NLP) research over the last years, offers novel business opportunities for companies, as automated user interaction or improved data analysis. Building sophisticated NLP applications requires dealing with modern machine learning (ML) technologies, which impedes enterprises from establishing successful NLP projects.
Our experience in applied NLP research projects shows that the continuous integration of research prototypes in production-like environments with quality assurance builds trust in the software and shows convenience and usefulness regarding the business goal. We introduce \gls{s4n} as an iterative and incremental process model for developing NLP applications.
With \gls{s4n}, we merge software engineering principles with best practices from data science. Instantiating our process model allows efficiently creating prototypes by utilizing templates, conventions, and implementations, enabling developers and data scientists to focus on the business goals. Due to our iterative-incremental approach, businesses can deploy an enhanced version of the prototype to their software environment after every iteration, maximizing potential business value and trust early and avoiding the cost of successful yet never deployed experiments.

\keywords{Natural Language Processing \and Process Model \and Machine Learning \and Best Practices \and Avoiding Pitfalls \and Quality Assurance.}
\end{abstract}
\section{Introduction}
The field of artificial intelligence in general and natural language processing as one of its sub-fields offers tremendous novel business opportunities in a steadily growing market \cite{stat-glob}. Recent progress in NLP research shows the potential for business applications, leading to a demand for more advanced NLP applications \cite{devlin-etal-2019-bert}.

The state-of-the-art in NLP differs from research to industrial domains. Besides the progress in research, the application of ML-based NLP in many enterprises is severely limited \cite{gartner-ratio-ai}. The black-box behavior of ML models, missing know-how, complex technological landscape, and the decision on an appropriate tool stack discourage enterprises from implementing NLP approaches \cite{xai,laurence-goasduff-2019}. They discard promising projects due to the combination of high and uncertain effort estimation \cite{staff-2019}. 

Many ML projects fail because of exceeding budgets, deadlines or they do not meet the business requirements \cite{failing-projects}. The late integration of several projects can lead to a services shutdown \cite{ms-chatbot,amazon}. We minimize the risk of these situations with agile methodology to handle the uncertainties and generate business value and feedback on the application as early as possible. This increases the quality and trust in the software for all involved stakeholders \cite{cohn}. 

We propose a new process model adjusted for developing NLP applications:
\textbf{Sta}ndardized \textbf{M}odeling \textbf{P}rocess for \textbf{N}atural \textbf{L}anguage \textbf{P}rocessing  (\gls{s4n}).

With \gls{s4n}, we merge software engineering principles with best practices from data science to improve and accelerate the development cycle and integrate prototypes with every iteration into a test or production environment. \gls{s4n} provides a transparent development process, including roles, tasks, artifacts, and best practices. 

Our main contributions\footnote{\url{https://github.com/philipp-kohl/stamp4nlp}} are:
\begin{itemize}
    \item A novel process model for developing NLP applications, with formally specified roles, activities, and artifacts focusing on quality, and early business value.
    \item Usage of predefined environment and software templates based on prior experiences for accelerating the development start.
\end{itemize}

\section{Related Work}
\label{sec:related-work}

Knowledge Discovery in Databases (KDD) \cite{fayyad-data-mining-kdd-1996} represents one of the first process models for data mining. It offers a generic guided process of the technical tasks to reveal patterns in data and building knowledge. 

Cross Industry Standard Process for Data Mining (CRISP-DM) \cite{crispdm} also considers business requirements and models the application development process in an applied context in contrast to KDD.
Modern ML process models originate from CRISP-DM. It consists of six stages: Business Understanding, Data Understanding, Data Preparation, Modeling, Evaluation, and Deployment. Depending on the stage's results, it allows transitions to previous stages. 

CRISP-ML(Q) \cite{crispmlq} extends CRISP-DM for machine learning and explicitly considers the differences between data mining (revealing patterns in data) and machine learning (training and inference). Studer et al. focus on quality assurance on every specific task. CRISP-ML(Q) merges Business Understanding and Data Understanding into a single stage and adds the Monitoring and Maintenance stage, addressing particular challenges of machine learning applications not considered by CRISP-DM. 

While CRISP-DM and CRISP-ML(Q) mainly focus on application creation, Weber et al. \cite{weber} introduced an approach with defined transitions between model development and model operation. Thus, they cover the whole model lifecycle from planning over production until retirement. Their process does not explicitly consider business requirements. 

Similar to KDD, Amershi et al. \cite{amershi} focuses on the modern, mainly technical process for developing a machine learning model but also incorporates fundamental operational analytics with transitions to previous stages.

In contrast to the mentioned process models, we focus on NLP. We treat manual data annotations and annotation guidelines as central project artifacts. Our approach is especially suitable for supervised NLP tasks as information extraction. Further, our focus is on strong quality assurance with different levels of applied tests. We leverage approaches and best practices from software engineering, combining them with machine learning approaches, such as supporting versioning of code, data, models, and tracking experiment results \cite{pineau_online}.

Agile software development \cite{cohn} uses iterations and increments instead of transitions between process stages. This leads to a stronger focus on running software during the development process. The incremental aspect allows isolated investigation of the experimental effects and the usage of continuous integration and delivery (CI/CD) \cite{ci-book}. Developers receive feedback, and stakeholders gain value and trust in the application with every deployed increment. In comparison to \cite{crispmlq} with task-agnostic quality assurance, we measure technical and business-oriented metrics on a higher level after every iteration \cite{checklist,mediation,attestation}.

Inspired by Spring \cite{spring}, maven archetypes \cite{maven-archetypes} and NLPf \cite{nlpf}, we deliver \gls{s4n} with a framework, which supports the developer with a predefined development environment, code for well-known standard tasks \cite{ner-survey,cl-survey} for creating a rapid prototype as basline or proof of concept, which the developer enhances over iterations and increments. \gls{s4n} decreases the risk of not deployed valuable experiments and failing projects.

\section{Process Model}
\label{sec:process-model}
\gls{s4n} is an instantiable, iterative, and incremental process model for developing natural language processing applications with a focus on quality, business value, and simplified prototyping.

\begin{wrapfigure}[23]{R}{8cm}
\includegraphics[width=8cm, keepaspectratio]{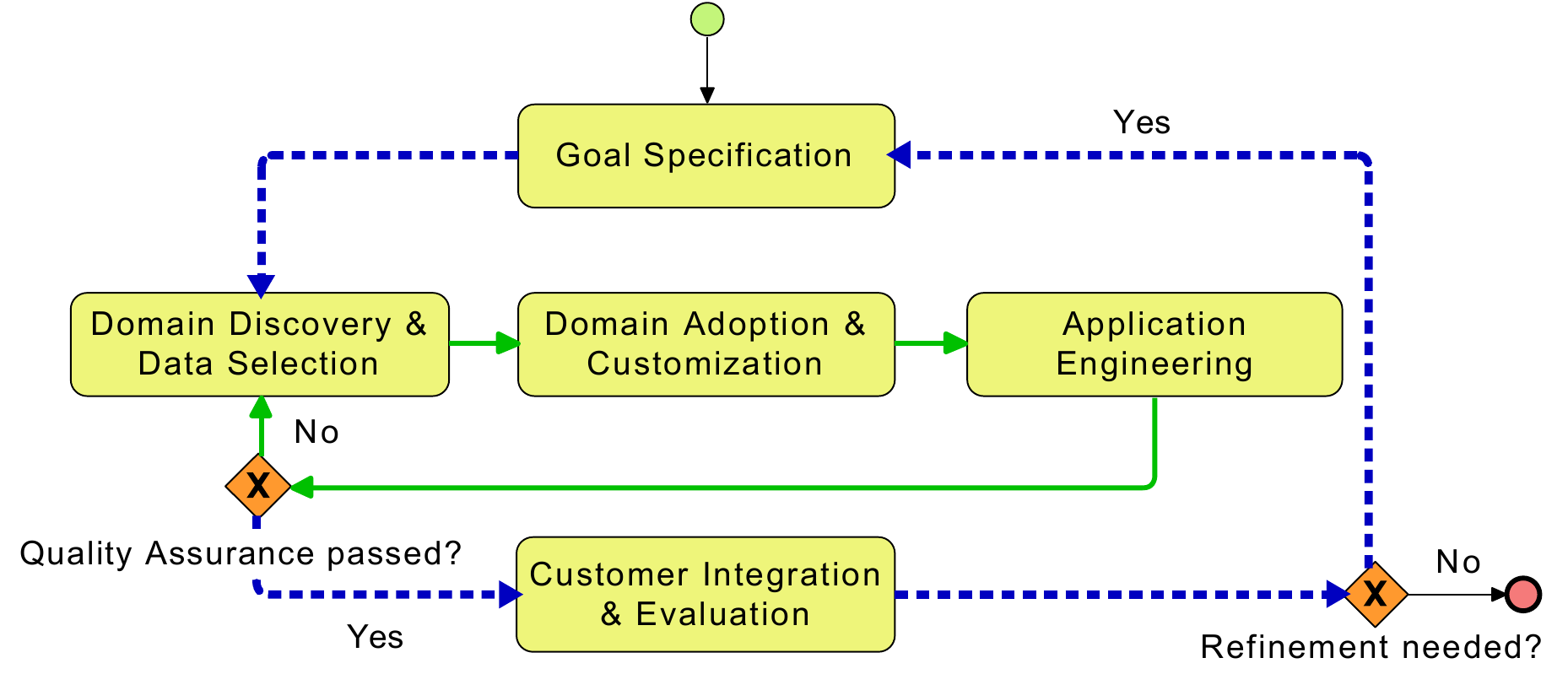}
\caption{\gls{s4n} consists of five subprocesses arraned in two nested loops: \textit{Development Loop} (green + solid) for developing the NLP application and model. \textit{Evolution Loop} (blue + dashed) surrounds the Development Loop and covers mainly the interaction with the customer: creation and refinement of the project goals with associated requirements and integration and monitoring of the application. Note: The solid and dashed connectors have the same meaning of sequence flow but help distinguish the loops in black and white print.}	
\label{fig:stamp-overview}	
\end{wrapfigure}

\gls{s4n} uses agile methodology \cite{cohn} by establishing software in increments developers enhance in various iterations (c.f. \autoref{fig:stamp-overview}).  
The Evolution Loop works in sprints (e.g., 2-4 weeks) to create a new release candidate that the developers integrate into the customer's test or production environment to gain trust in the software and receive feedback for further improvements to fit the customer's needs \cite{ci-book}. The workflow is as follows: the developers start with the Evolution Loop to define or refine the requirements, followed by multiple iterations in the Development Loop (hours to days) to create a model that satisfies the specified requirements. Deployment and testing are also a part of the inner loop. Once the software fulfills them, the developers leave the inner loop and proceed to the outer loop to integrate and monitor the release candidate. Depending on the monitoring results, they trigger a new iteration.

The process model provides particular levels of transparency. At first, every stakeholder can follow the complex process, even if they are unfamiliar with NLP. They can also track the order of tasks by consulting the documentation. Thereby the responsibilities are clearly defined, and every involved party knows their tasks. Clearly defined responsibilities are helpful to prevent conflicts and create rational processes that the users can monitor and optimize. We use the\textit{MEDIATION} \cite{mediation} approach to provide the application state transparent via a dashboard with project specific business metrics for all stakeholders.

The development environment equips the user with a standard set of tools for rapidly building a proof of concept while maintaining the flexibility to let the user choose other preferred libraries.
Creating a project instance provides folder structure, development environment, documentation, and boilerplate code (e.g., REST-Service with predefined endpoints, etc.). The standard folder facilitates the automatic loading and storing of data, models, and results to their destination by convention. Furthermore, no settling-in period for already STAMP-involved members; they know where to find code, documentation, experiments, and data.

ML metrics serve as a common benchmark for different models. High or low metrics show the averaged performance but do not allow making conclusions which use-cases a model cannot handle appropriately. Model interpretability is a current research subject \cite{xai}. 
To minimize the interpretability gap in machine learning and to receive more feedback on a low application level, we incorporated CheckList \cite{checklist} for creating behavioral tests aiming for specific capabilities the NLP application should cover.
Besides CheckList, we use MEDIATION \cite{mediation} for testing the NLP application on a business level: the developers and stakeholders define test cases strongly related to their intended use cases in the form of annotated documents. Thus, this additional test set serves as an indicator of the business readiness of the NLP application.
In combination with the CI/CD approach, we receive this feedback for every iteration and increment. Involving testers or real user groups into the increment testing generates feedback for business and practical usage.

\textit{\gls{s4n}} supports the user to keep the reproducibility of experiments and models as high as possible. We incorporated parts of Pineau's reproducibility list \cite{pineau_online} into the process by documentation and tools supporting versioning of code, data, models, and tracking experiment's results.

In the following, we give a short description of each subprocess with its primary artifacts. We show exemplarily a detailed BPMN diagram of the Domain Adoption and Customization in \autoref{fig:domain-adopt}. Our GitHub repository provides the other subprocess diagrams and the detailed description of each task and artifact.

\subsection{Goal Specification}
\label{sec:goal-spec}

\textbf{Description:} The Goal Specification aims to establish a common business understanding. The stakeholders define and refine the business goals, and their associated technical, machine learning, business, and MEDIATION requirements and update the documentation accordingly. It includes an evaluation of all data sources and the data provision for the data scientists. This stage involves all currently relevant stakeholders to minimize the bias and possibly wrong model assumptions.

\noindent \textbf{Artifacts:} The primary artifacts are the refined and reviewed requirements, test cases, and access to all mandatory data sources.

\subsection{Domain Discovery and Data Selection}
\label{sec:domain-disc}

\begin{figure}
    \centering
	\includegraphics[width=0.93\linewidth]{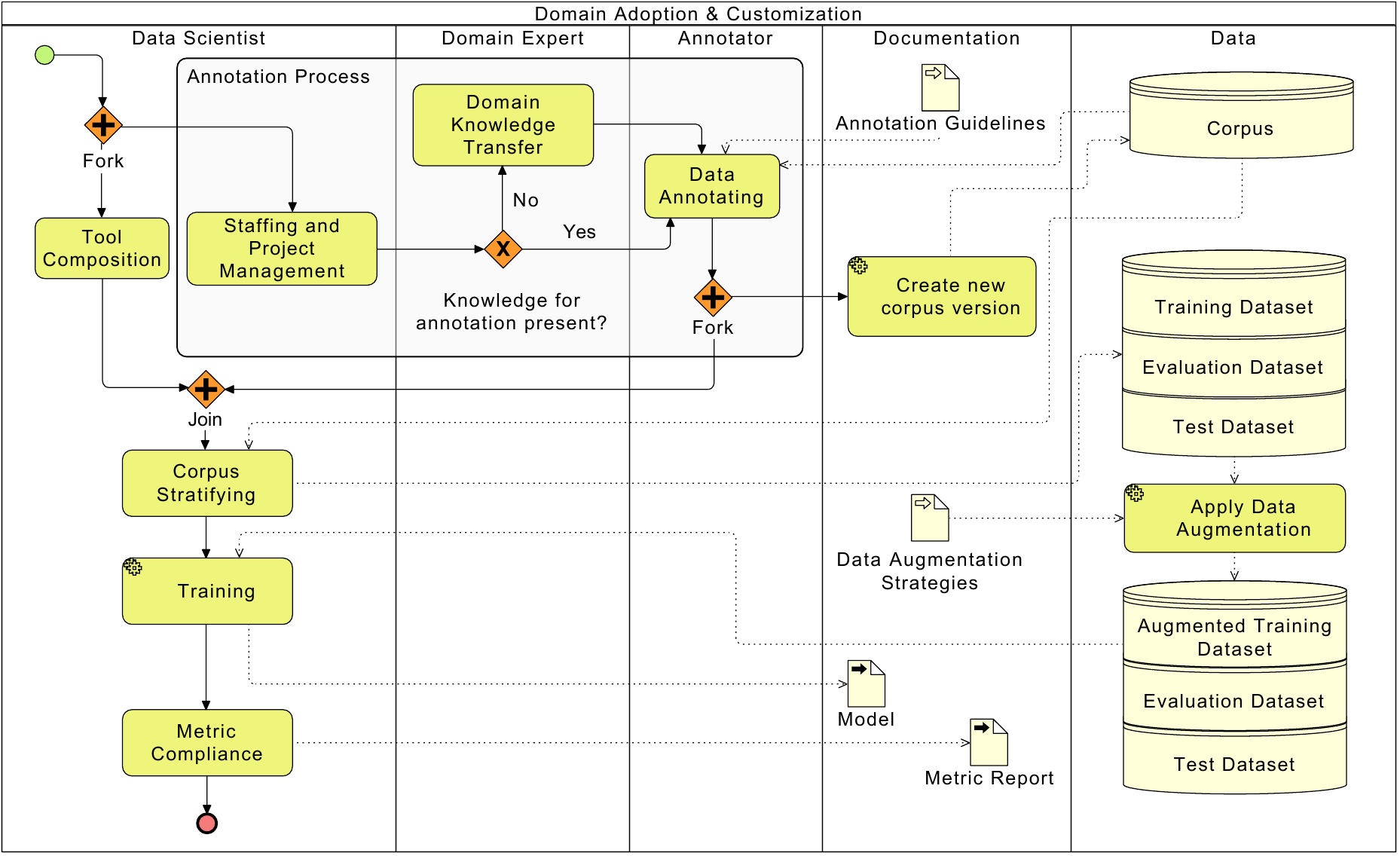}
	\caption{Showing the task order of the subprocess \textit{Domain Adoption and Customization} as BPMN Diagram with the involved roles displayed as swimlanes. Additional information is provided by \textit{Documentation} and \textit{Data} lane. \textit{Create new corpus version}, \textit{Training}, and \textit{Apply Data Augmentation} are automatic tasks, which run without human interaction. \textit{Training} requires the execution of one command as trigger.}
	\label{fig:domain-adopt}	
\end{figure}

\textbf{Description:} Data scientists and domain experts prepare the annotation process (c.f. \autoref{fig:domain-adopt}). They identify NLP tasks and corresponding annotation schemas helpful to fulfill the business goals. Additionally, data scientists include domain knowledge from experts to steadily improve annotation guidelines and collect and evaluate data samples for the annotation process and necessary metadata.

\noindent \textbf{Artifacts:} The primary artifacts are the annotation guidelines, the new corpus versions prepared for annotation, and documentation about licenses, data protection, and data security.

\subsection{Domain Adoption and Customization}
\label{sec:domain-adopt}

\textbf{Description:} This subprocess (c.f. \autoref{fig:domain-adopt}) includes data annotation and model training. The annotation process setup involves planning and, if necessary a domain training for annotators. The annotated texts build a new corpus version, which is used for training a new model.
To minimize the annotation effort, we want to stop further annotating when noticing the resulting model's metrics stagnate. We incorporated the continuous integration and delivery approach of Schreiber et al. \cite{cicd}. Thus, annotators receive feedback after each annotation session. The feedback can motivate to continue or stop annotating because the further annotations do not impact the model's performance remarkably.

\noindent \textbf{Artifacts:} The primary artifacts are the new corpus version, the new model, and metric reports about the model's performance.

\subsection{Application Engineering}
\label{sec:application-engi}

\textbf{Description:} Software developers package the model and all necessary dependencies. A CI/CD pipeline deploys the application in a test or production-like environment and runs evaluations to ensure the quality gates (software quality, machine learning metrics MEDIATION and behavioral tests via CheckList \cite{checklist}) defined in the goal specification. The software package is versioned to provide transparency, whether the made modifications improved the previous version and for fallback solutions. Depending on the evaluation results, the application stays in the Development Loop or transitions into the Evolution Loop for integrating the software into the customer's application landscape (c.f. \autoref{fig:stamp-overview}).

\noindent \textbf{Artifacts:} The primary artifacts are the software package and a quality assurance report.

\subsection{Customer Integration and Evaluation}
\label{sec:customer-int}

\textbf{Description:} Software developers integrate the packaged NLP application in the customer's application landscape. On the customer side, a monitoring service checks the model's performance. The resulting reports build the basis for a refinement of the quality gates comparable to \cite{attestation,mediation} ensuring the fulfillment of the business requirements during the production phase.

\noindent \textbf{Artifacts:} An operation manual documents the deployment, and an integration plan explains the integration in the customer application. The performance reports support recommendations and business decisions.

\section{Project Template}
\label{sec:implementation}
\gls{s4n} facilitates focusing on the project-specific challenges such as analyzing the data and the domain, annotating, experimenting with different concepts, and deep learning architectures.
To decrease the overhead data scientists face while starting a new project, we offer a template with development environment, folder structure\footnote{similar to \url{https://drivendata.github.io/cookiecutter-data-science/}}, tools, code and process documentation.
Furthermore, the framework can generate customizable implementations for specific NLP tasks (e.g., named entity recognition (NER) \cite{ner-survey}, or text classification \cite{cl-survey}) into the project, helping the developers implement a first baseline or proof of concept.

The template serves the paradigm \textit{convention over configuration} \cite{conv-over-conf}. Therefore the template comes with, but is not limited to a standard set of tools and libraries. If the user stays with the standard, no additional configuration is needed. But the user has the opportunity to use additional libraries or tools, resulting in extra configuration. The basic configuration provides, for example, the library spacy\footnote{\url{https://spacy.io/}} to create prototypes quickly. Depending on the business goal, it is necessary to preserve more control over used architecture and training routines. Therefore the developer can exchange the conventional added spacy module with PyTorch\footnote{\url{https://pytorch.org/}} or similar frameworks. The same applies to the folder structure, environment, and infrastructure. We recommend to start with the standard configuration and specialize on demand.

\section{Example}
\label{sec:loop-example}

This section demonstrates a simplified \gls{s4n} usage over a few iterations to show the intuition behind the process model. We focus on a real-world project we performed with our business partners: The profile extraction from social media messages of an advertising group conversation. We use named entity recognition (NER) as a standard NLP task, for which we can use existing approaches. NER describes the task of finding domain-relevant terms in documents: e.g., persons, brands, products, and their prices. On top of that, we implement a business layer to aggregate the named entities to a profile.

\textbf{First iteration -- Requirements Analysis and Dry Run:} 
The first iteration focuses on the requirements analysis and the infrastructure test run (also called \textit{dry run}). Instantiating the process model provides a development environment including a prototypical web-service, a pre-configured CI/CD pipeline, and prepared documentation. We define and document the NER as the applied NLP task. Based on the documentation, we invoke the framework for generating a reference implementation for NER as a first baseline. Among others, we define the corresponding machine learning metrics we want to achieve with the NLP application. 
We skip the most tasks of all other subprocesses for this iteration since its the iteration's goal to ensure the infrastructure: training a model, embed the model into a software package, deploy into a test environment, model evaluation, publish results via a dashboard. The scores do not matter at this stage.
        
\begin{wrapfigure}[24]{R}{6.5cm}
    \includegraphics[width=\linewidth]{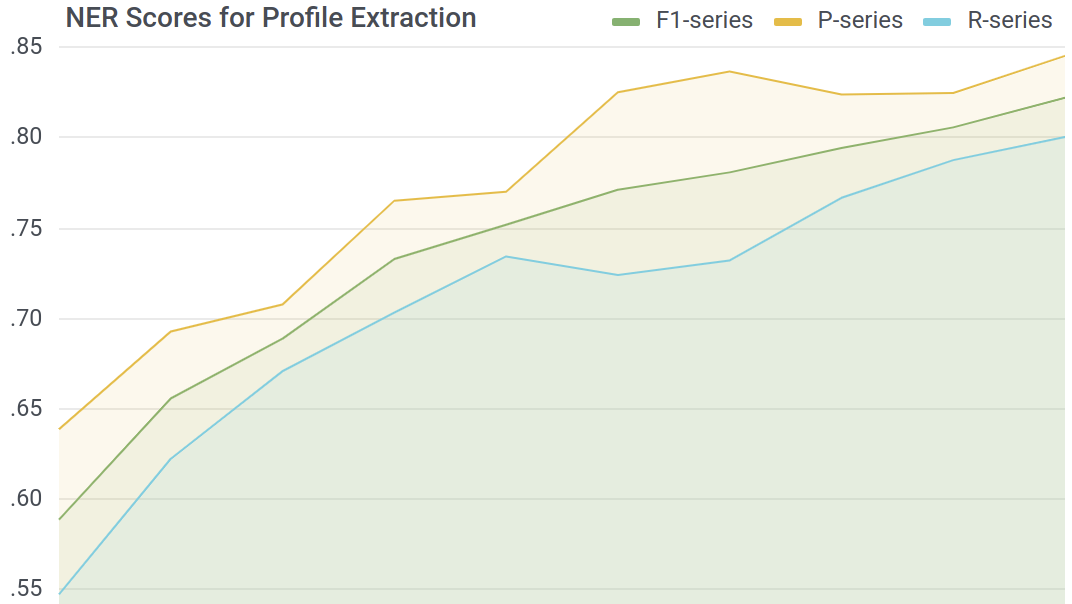}
    \caption{An excerpt of a Grafana Dashboard visualizes the F1, precision, and recall score for the NER component over several iterations of annotating and architecture decision. Precision measures the correctness of positive classifications (here for a named entity) by penalizing false positives. Recall shows if our model has found all named entities by penalizing false negatives. F1 is the harmonic mean of precision and recall. It is commonly used for model evaluation and comparison.}	
    \label{fig:dashboard}	
\end{wrapfigure}

\textbf{Second iteration -- Baseline:}
The second iteration focuses on creating a first baseline. This iteration covers the \textit{Development Loop} exclusively.
We prepare the annotation process by defining the annotation guidelines, deciding on a suitable corpus format, and transferring the data to the corpus format in Domain Discovery and Data Selection.
Domain Adoption and Customization mainly focuses on annotating a subset of the data in this iteration to create the first baseline with the standard implementation. 
We embed the model into our software architecture and add a business layer to combine the named entities to a profile. Our CI/CD approach packages and deploys the software into the test environment, where the framework performs detailed quality assurance. The results are published on a dashboard (c.f. \autoref{fig:dashboard}).

\textbf{Additional iterations -- Beat the baseline:}
Further iterations focus isolated on specific aspects to enhance the baseline. We incorporate the aspects as isolated as possible to ensure the cause-effect relationships:
\begin{description}
    \item[Annotations] Annotating new data points, improve existing annotations, decide to incorporate new labels, enhance annotation guideline based on gained experience.
    \item[Architectural Decisions] Use different common neural network components or pretrained contextualizes embedding layers \cite{elmo,devlin-etal-2019-bert}.
\end{description}
If the application fulfills the requirements, we have a possible release candidate and exit the Development Loop and continue the Evolution Loop.
If the results do not show the expected behavior or do not fulfill the business need, we begin a new iteration to investigate the cause and start new experiments improving the application's quality.

\section{Limitations and Drawbacks}

Our proposed process model has a strong focus on supervised NLP. We have many subprocesses with corresponding roles and artifacts for problems that include the manual annotation of corpora. They may be unbeneficial for unsupervised tasks. We further define clear responsibilities and processes but assume that a practical application involves loose compliance to those in some cases. The process needs to be adapted to each project individually. A significant benefit of the instantiable framework is a low application barrier. The resulting standardized configuration and black box code can lead to laborious error detection or adaptions when moving too far from these conventions.

\section{Summary}
\label{sec:summary}

We introduced \gls{s4n} a novel instantiable and iterative-incremental process model to develop NLP applications. It supports developers to create valuable and deployable increments rapidly, results in earlier feedback, and improves quality and trust in the application for all stakeholders. 

Our approach equips the user with templates, development environment, and documentation to reduce the starting and integration overhead. That minimizes implementation barriers, avoids common pitfalls, and sets the focus on the business goal. Thus, \gls{s4n} reduces the risk of failing projects.

In the future, we plan to create a benchmark project for different groups to work on: some groups work with the \gls{s4n} and others work from scratch. Thus, we want to measure various project milestones, key performance indicators and observe the challenges the different teams face and pitfalls they could avoid. 
Furthermore, we want to use it for educational purposes to set the focus appropriately with incremental depth increase. We want to improve the development of generic NLP tasks, including unsupervised problem settings.

\bibliographystyle{splncs04}
\bibliography{main}
\end{document}